# A Comparative study of Arabic handwritten characters invariant feature

Hamdi Hassen
Mir@cl Lab, FSEGS
University of Sfax
BP 1088, 3018Sfax, Tunisia
(216) 74 278 777
hhassen2006@yahoo.fr

Maher Khemakhem
Mir@cl Lab, FSEGS
University of Sfax
BP 1088, 3018 Sfax, Tunisia
(216) 74 278 777
maher.khemakhem@fsegs.rnu.tn

*Abstract*—this paper is practically interested in the unchangeable feature of Arabic handwritten character. It presents results of comparative study achieved on certain features extraction techniques of handwritten character, based on Hough transform, Fourier transform, Wavelet transform and Gabor Filter. Obtained results show that Hough Transform and Gabor filter are insensible to the rotation and translation, Fourier Transform is sensible to the rotation but insensible to the translation, in contrast to Hough Transform and Gabor filter, Wavelets Transform is sensitive to the rotation as well as to the translation.

## I. INTRODUCTION

In order to improve the recognition rate of Arabic handwritten character, several techniques based on geometric correction or mathematic transform have been tested and applied on the input image of such writing. The purpose of all these techniques is to attempt to find the invariant features of Arabic handwritten characters. Unfortunately, all these attempts failed because, researchers didn't try to take advantages of each one of these techniques and then try to build an approach based on the cooperation of some (or all) of them.

Consequently, we show in this paper the behavior of some well-known mathematic transforms to decide in a next step on which of them can be used especially in a future Arabic OCR system based on the cooperation of several approaches and techniques.

The remainder of this paper is organized as follows: the first section presents a detailed description of the bench of test used (retained system) and the corresponding features. The second section, gives an assessment of this system, whereas the last section compares and analyses the obtained results.

## II. THE RETAINED SYSTEM

Due to the nature of handwriting with its high degree of variability and imprecision obtaining these features, is a difficult task [1]. Feature extraction methods are based on 3 types of features: statistical, structural and global transformations and moments.

Several mathematic transform have been adapted to the Arabic character as well as for the modeling of the different morphological variations of the characters and for the resolution of the difficult problems of the segmentation.

### A. Hough Transform

The Hough transform is a very general technique for feature detection. In the present context, we will use it for the detection of straight lines as contour descriptors in edge point arrays [2].

The Hough transformation (HT) is proposed since 1962 in US by P. V. C. Hough as the patent, because it may carry on the effective recognition to the shape, and parallel realizes, moreover is insensitive to the noise, thus obtained the enormous attention. In the practical application, mainly uses in recognizing curve[3].

Hough transform is a methodology which consists of the following three steps; the first one includes preprocessing for image enhancement, connected component extraction and average character height estimation. In the second step, a block-based Hough transform is used for the detection of potential text lines while a third step is used to correct possible false alarms [4].

#### 1) Preprocessing

First, an adaptive binarization and image enhancement technique is applied [5]. Then, the connected components of the binary image are extracted [6], and for every connected component, the bounding box coordinates and the corresponding area are calculated [7]. Finally, the average character height $AH$ for the document image is calculated [6]. We assume that the average character height equals to the average character width $AW$.

#### 2) Hough Transform Mapping

In this stage, the Hough transform takes into consideration a subset (denoted as "subset 1" in Figure 1) of the connected components of the image. This subset is chosen for the following reasons: (i) it is required to ensure that components which appear in more than one line will not vote in the Hough domain; (ii) components, such as vowel accents, which have a small size must be rejected from this stage because they can cause a false text line detection by connecting all the vowel above to the core text line.

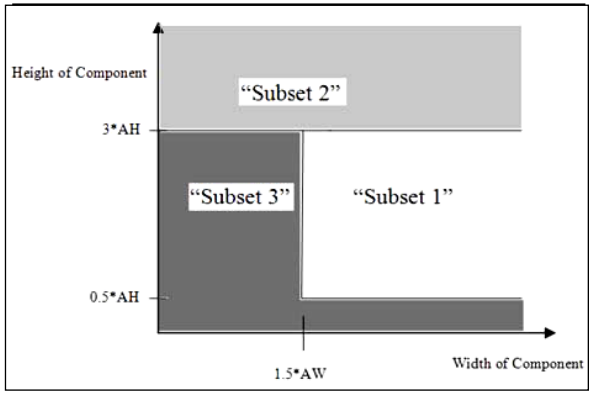

**Figure1.** The connected component space divided into 3 subsets denoted as "Subset 1", "Subset 2" and "Subset 3".

The spatial domain for "subset 1" includes all components with a size identified by the following constraints [6]:

$$0.5*AH < H < 3*AH$$

$$1.5*A\Omega < \Omega$$

Where *H, W* denote the component's height and width, respectively, and *AH, AW* denote the average character height and the average character width, respectively.

After this partitioning stage, we calculate the gravity center of the connected component contained in each block which is then used in the voting procedure of the Hough transform.

The Hough transform is a line to point transformation from the Cartesian space to the polar coordinate space. Since a line in the Cartesian coordinate space is described by the equation [8]:

$$P = X \cos(\theta) + Y \sin(\theta) \qquad [1]$$

To construct the Hough domain the resolution along *θ* direction was set to 1 degree letting *θ* take values in the range 85 to 95 degrees and the resolution along *p* direction was set to 0.2*AH* [9]. After the computation of the accumulator array we proceed to the following procedure: we detect the cell ($p_i$, $\theta_i$) having the maximum contribution and we assign to the text line ($p_i$, $\theta$ i).all points that vote in the area ($p_i$ -5, $\theta$ i) .. ($p_i$ +5, $\theta$ i)

*3) 3) Post processing*

The previous stage may result in more than one line assigned in the Hough domain that corresponds to a single text line (see Figure 1).

This correspondence is determined by calculating the distance between the corresponding crossing points of the lines with the document middle vertical line [10].

If the distance is less than the average distance of adjacent lines then all connected components which correspond to these lines are assigned to the same text line label. In this step we determinate the subset 2 and 3 (figure 1) "Subset 2" includes the components whose heights exceed 3 times the average height [6] (see Figure 1). These 'large' components may belong to more than one text line. This situation may appear when an ascender of one line meets a descender of an adjacent line. "Subset 3" includes all the components that do not fall into the previous two categories. Components of "Subset 3" are usually punctuation marks or elongations.

*B. Wavelet Transform*

Wavelet Transform is the method of decomposition that has gained a great deal of popularity in recent years. It seeks to represent a signal with good resolution in both time and frequency, by using basis functions called wavelets.

This algorithm consists in the following: for every level of decomposition, we make the extraction of the average of vertical, horizontal, right and left, directional and global features. Then, we decompose the picture of size 512 *512 in blocks of size 4*4 stamps to have in total 16 * 5=80 features then one calculates the black pixel density for every block [11]. The related components that have an elevated frequency density are filtered by 2D wavelets.

Mathematically, the process of Wavelets Transform is represented by the following equation:

$$F(w) = \int_{-\infty}^{+\infty} f(t) e^{-jwt} dt \qquad [3]$$

With

F(w): Wavelet Transform result,

$f$(t) : Input picture,

W: Sinusoïde of frequency.

Wavelet transform are of different types: Haar Wavelet Transform, Symlet Wavelet Transfor, Daubechies (db4)Wavelet transfor, ..[12]

For every type, we can have two types of wavelet transform: the Continuous Wavelet Transform (CWT) and the Discrete Wavelet Transform (DWT) [13]. Moreover the wavelet is classified in several families Haar, Daubechies, Symlets, Coiflets…

And For every category, we can have several levels of cœfficients [14] Daubechies (db3), Symlets (sym5), Coiflet2,… of which every category uses a cœfficient that gives an invariant average in several representation  to different resolution.

*C. Gabor Filter*

Gabor Filter can capture salient visual properties such as spatial localization, selectivity orientation, and spatial frequency characteristics [15][16]. We have chosen Gabor features to represent the face image considering these excellent capacities [17]. Gabor filters are defined as follows [18]:

$$G(x,y,\theta,f) = \exp\left(\frac{-1}{2}\left(\frac{x'}{sx'}\right)^2 + \left(\frac{y'}{sy'}\right)^2\right) * \cos(2\pi * f * x') \quad [2]$$

With

$x' = x*\cos(\theta) + y*\sin(\theta)$

$y' = y*\cos(\theta) - x*\sin(\theta);$

I: Input picture

Sx & Sy: The Variances of the x and y axes respectively

f: the frequency of the sinisouide function

$\theta$: The orientation of Gabor filter

G: the output of Gabor filter.

The feature vector used by the OCR system consists in the combination of several parameters such as $\theta$, f and the other parameters [18].

The rotation of the characters in the X and Y axes by angle $\theta$, results in a vector of feature with an orientation $\theta$.

The value $\theta = \prod (k-1)/m$ with k= 1… m and m is the number of orientations

The position of a character in a N*N picture is taken into consideration in order to get a result of invariant feature vector to the rotation and translation [19].

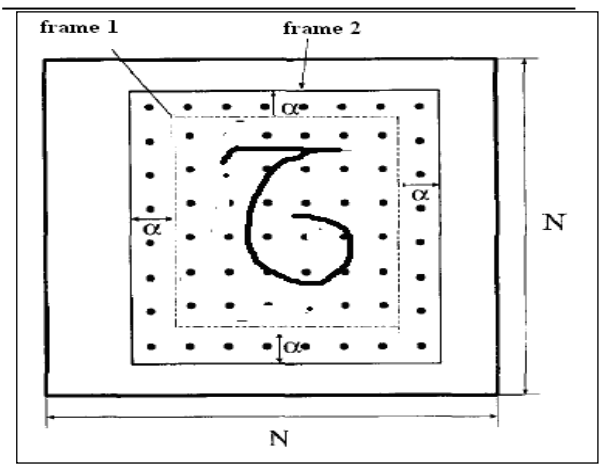

Figure2. Position of the character «Haa» in a picture of [N*N] size

*D. Fourier Transform*

The Fourier Transform (FT) of the contour of the image is calculated. Since the first n coefficients of the FT can be used in order to reconstruct the contour, then these n coefficients are considered to be a n-dimesional feature vector that represents the character

The recognition of the writing by Fourier Transform is operated previously on the contour of the character [20]. At the first stage, we start with the detection of the contour. In the second stage, the code of Freeman of the contour is generated on which one operates the calculation of Fourier Transformed.

The figure below shows the stages of character recognition by Fourier Transform.

The method Fourier Transform requires the modelling of the function of the contour. The contour of a character is described by a sequence of the codes of Freeman characterized by [20]:

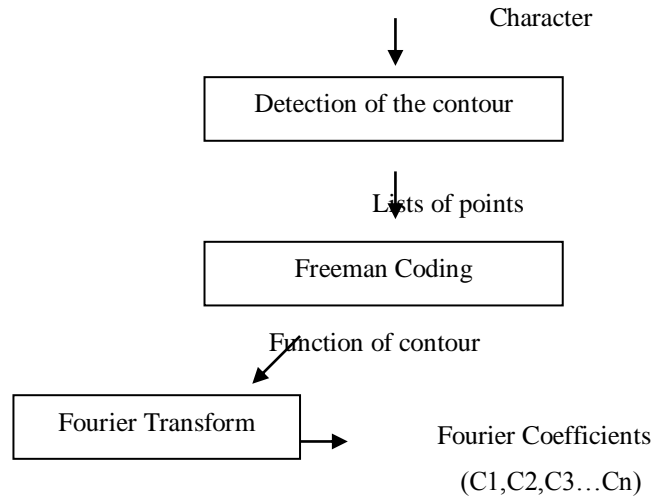

Figure 3. The stages of Fourier Transform

The values: between 0 and 7,

The direction: of 0 to $2\pi$ per step of $\pi/4$ while following the opposite direction of the needle of a watch,

The length: 1 or $\sqrt{2}$ according to the parity of the codes.

From this description, we can generate a periodic temporal function on which Fourier Transform can be calculated.

Mathematically, the process of Fourier Transform is represented by the following equation:

$$X_N(K) = A_0 + \sum_{n=1}^{N} a_n \cos\frac{2\pi nK}{N} + b_n \sin\frac{2\pi nK}{N} \quad [4]$$

$$Y_N(K) = C_0 + \sum_{n=1}^{N} c_n \cos\frac{2\pi nK}{N} + d_n \sin\frac{2\pi nK}{N} \quad [5]$$

Where

K: the kème points of the contour,

N: is the necessary number in the approximation of the contour by the coefficients of Fourier

$a_n$, $b_n$, $c_n$ et $d_n$: The coefficients of Fourier corresponding to the harmonic n.

$a_0$ et $c_0$: The continuous components that correspond to the initial points where the frequency is equal to 0.

## III. EVALUATION OF THE RETAINED SYSTEM.

### A. Implementation

For some experimental convenient, the MATLAB Version 7.4.0 (R2007a) (MATrix LABoratory) has been used.

### B. The IFN/ENIT Handwritten Character Database

We have used the well known IFN/ENIT corpus data base formed of handwritten Tunisian town's names.

The figure 4 presents a sample basis of the IFN/ENIT.

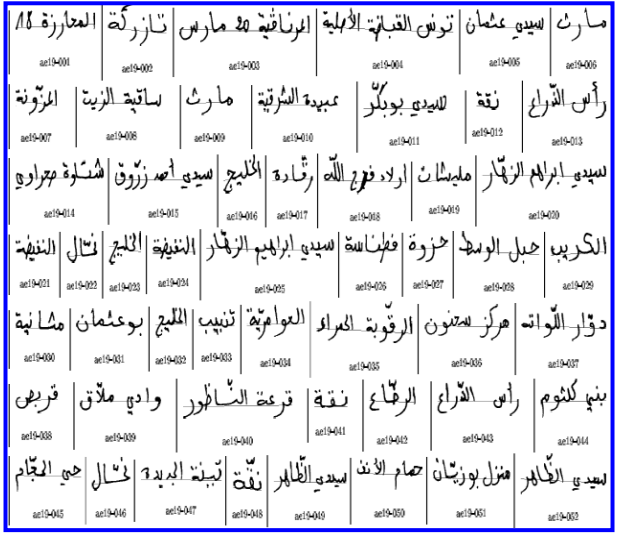

Figure 4. Some names of Arabic Tunisian towns of the IFN/ENIT

These names have undergone of the segmentations, the final version of the retained basis includes the different handwritten Arabic characters in the different positions (initial, median, isolated).

We note that the set of characters, the signs of punctuation as well as the numbers are written by different writers. (The basis of training includes 692 pictures of letters whereas the basis of test includes 1674 which 692 those of the training and 982 new pictures of characters to be recognized :)

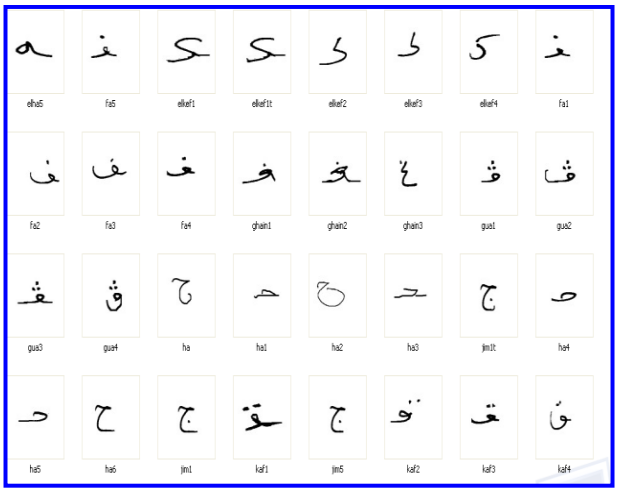

Figure 5. Some handwritten Arabic characters pictures

### C. The classification by Euclidean Distance

There are several methods of classification such as k-Nearest Neighbour (k-NN), Bayes Classifier, Neural Networks (NN), Hidden Markov Models (HMM), Support Vector Machines (SVM), etc [21].

There is no such thing as the "best classifier". The use of classifier depends on many factors, such as available training set, number of free parameters etc.

In our case, we used Euclidean Minimum Distance Classifier (EMDC).

The consistent rule is relative to the decision of the nearest neighbor's; the idea is extremely simple the character feature vector $x_i$ is compared with the vectors $y_i$ describing the character in the Base Set to search for the nearest neighbour of the tested character.

The general equation of the Euclidean distance is the following:

$$d(x,y) = \sqrt{\sum_{i=1}^{N}(x_i - y_i)^2} \quad [6]$$

### D. Test and evaluation

The implementation of our system requires the definition of a set of parameters:

Wavelets Daubechies (db) with a level of decomposition equal to 3 because they are used intensively and they present interesting properties.

FFT2: Fast Fourier Transform level 2 is the algorithm of Fourier used in our implementation because the calculation of the coefficients of discreet Fourier is less expensive.

The implementation of Hough Transform doesn't require a particular parameter; we simply program the different stages of this method.

The implementation of the Gabor Filters method requires the research of the most combinations of a set of parameters ($\lambda, \sigma_x, \sigma_y, m$) that maximizes the recognition rate. $m = 4$; $\lambda = 1$; $\sigma_x = 2$; $\sigma_y = 1$;

The results of our study are reported on the following figures.

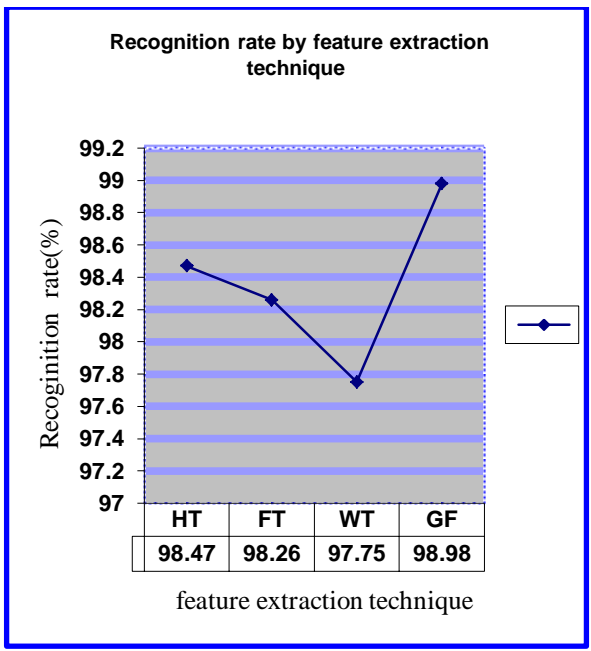

Figure 6. Recognition Rate by feature extraction technique

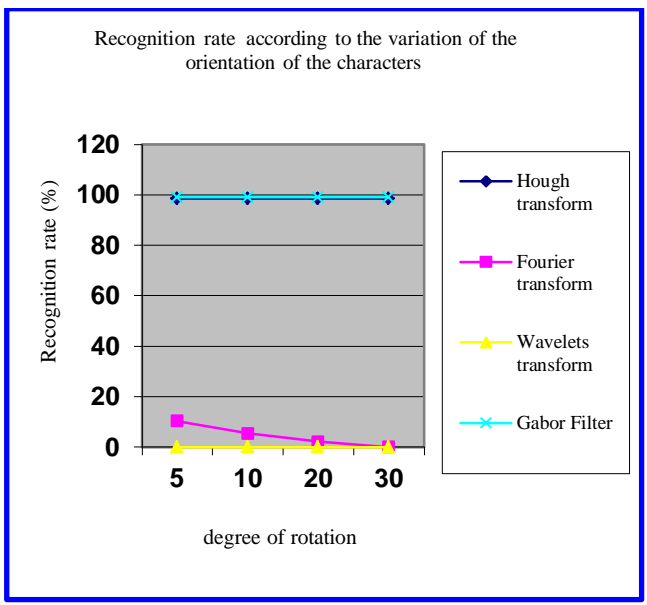

Figure 7. Recognition rate according to the variation of the orientation of the characters

Similarly, the transfer of the characters according to the x axis or y coordinates gives interesting results with Gabor, Hough and Fourier transforms.

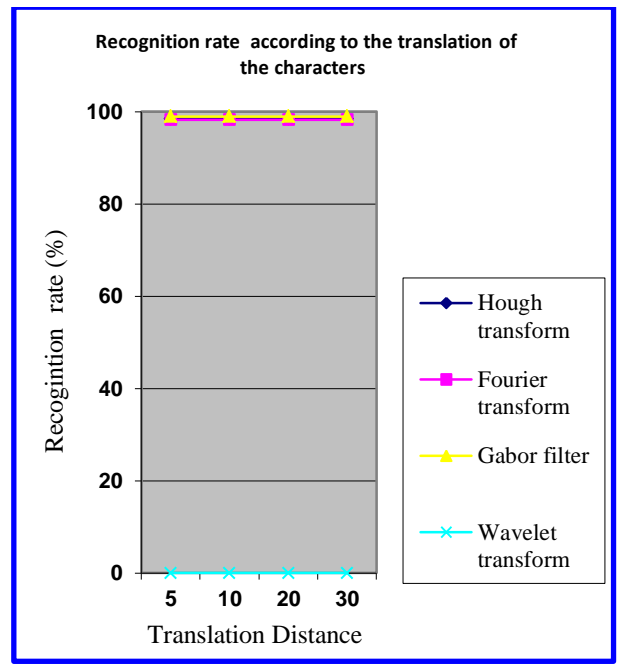

Figure 8. Recognition rate of according to the translation of the characters

After extensive experiments, we tried to calculate the response time of our recognition system.

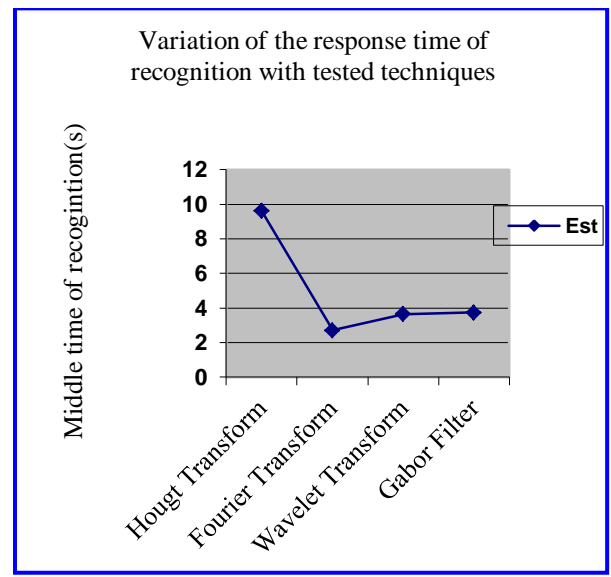

Figure 9. Variation of the response time of recognition with tested techniques

We have evaluated also the size characters (written) variation with the four extraction techniques of our system. Obtained results are reported in the figure below .

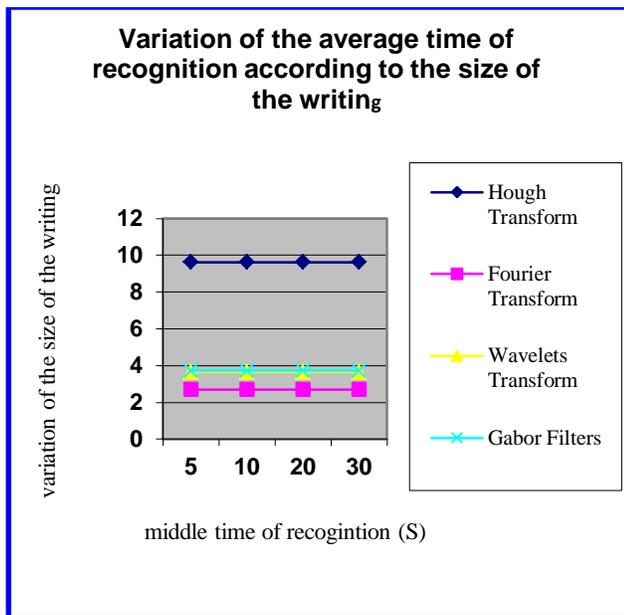

Figure 10. Variation of the average time of recognition according to the size of the writing

*E. Interpretation:*

The previous results show that:

The four feature extraction techniques can give interesting recognition rates depending to the initial conditions of the Arabic writing, Hough Transform and Gabor filter are insensible to the rotation and translation,

Hough Transform and Gabor filter are indeed efficient and strong tools for the detection of remarkable shapes in a picture,

Hough Transform, was very slow because of the corresponding complexity,

Fourier Transform is sensible to the rotation but insensible to the translation,

In contrast to Hough Transform and Gabor filter, Wavelets Transform is sensitive to the rotation as well as to the translation,

The writing size variation doesn't have an influence on the recognition average time for the four studied extraction techniques (average time is invariant)

## IV. SURVEY OF SOME DETECTED ERRORS.

The most common detected errors are the following

Confusion between the characters "gua" and « tha » ( Fourier transform),

Confusion between the characters « ya» and « hamza » ( Gabor Filter),

Confusion between the characters « kaf » and « fa » : the two diacritical points of the character " kaff " are confounded in one only. (Hough transform),

Confusion between the characters « elha » and « mim » (Wavelets Transform).

The two characters (mim) and (zed) have the same Euclidian distance (Wavelet transform),

The two characters (lam) and (del) have the same Euclidian distance (Hough transform)

## V. CONCLUSION AND FUTURE WORK

We present in this paper a comparative study of different features extraction techniques based on Hough Transform, Wavelet Transform, Fourier Transform and Gabor filters.

The obtained results provide interesting idea about the studied techniques.

These results show that Hough Transform and Gabor filter are insensible to the rotation and translation, Fourier Transform is sensible to the rotation but insensible to the translation, in contrast to Hough Transform and Gabor filter, Wavelets Transform is sensitive to the rotation as well as to the translation

The considered perspectives to make evolve the present work are numerous: we can note the evolution of other classification is one of them, exploiting new features to improve the current performance. The integration and cooperation of some complementary approaches that can lead to an effective Arabic OCR is also another way of investigation.

The digitization of the national cultural heritage is another future project or work

The project is expected to connect with vanguard digital libraries such as Google, and to digitize many books, periodicals and manuscripts.

Our project would make access to data much easier for researchers and students in all regions of the country.

We need a technology that offers a number of benefits, such as the ability to store and retrieve large amounts of data in any location at any time (Mobility of the users: dynamic environment).

ACKNOWLEDGMENT

The authors wish to thank the reviewers for their fruitful Comments. The authors also wish to acknowledge the members of the miracl laboratry, sfax, tunisia

AUTHORS PROFILE

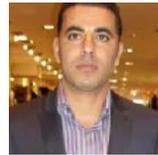
**Hassen Hamdi** received in 2008 his Master's Degree in Computer Science from the University of Sfax, Tunisia. He is currently a Ph.D student at the University of Sfax. His research interests include distributed systems, performance analysis, Networks security and pattern recognition.

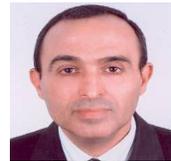
**Maher Khemakhem** received his Master of Science, his Ph.D. and Habilitation degrees from the University of Paris 11 (Orsay), France respectively in 1984, 1987 and the Universtity of Sfax, Tunisia in 2008. He is currently Associate Professor in Computer Science at the Higher Institute of Management at the University of Sousse, Tunisia. His research interests include distributed systems, performance analysis, Networks security and pattern recognition.